\def\year{2020}\relax
\documentclass[letterpaper]{article} 
\usepackage{aaai20}  
\usepackage{times}  
\usepackage{helvet} 
\usepackage{courier}  
\usepackage[hyphens]{url}  
\usepackage{graphicx} 
\usepackage{graphicx}  
\usepackage{amsmath}
\usepackage{amssymb}
\usepackage{amsthm}
\usepackage{arydshln}
\usepackage{listings}
\usepackage{subcaption}
\usepackage{tikz}
\usetikzlibrary{arrows,automata,shapes}
\newcommand{\tup}[1]{{\langle #1 \rangle}}

\newcommand{\pre}{\mathsf{pre}}     
\newcommand{\eff}{\mathsf{eff}}     
\newcommand{\cond}{\mathsf{cond}}   

\newtheorem{definition}{Definition}
\newtheorem{theorem}{Theorem}
\newtheorem{corollary}{Corollary}

\title{Generalized Planning with Positive and Negative Examples}

\author{\Large \textbf{Javier Segovia-Aguas\textsuperscript{\rm 1}, Sergio Jim\'enez\textsuperscript{\rm 2} and Anders Jonsson\textsuperscript{\rm 3}}\\ 
\textsuperscript{\rm 1}IRI - Institut de Rob\`otica i Inform\`atica Industrial, CSIC-UPC\\
\textsuperscript{\rm 2}VRAIN - Valencian Research Institute for Artificial Intelligence, Universitat Polit\`ecnica de Val\`encia \\
\textsuperscript{\rm 3}Universitat Pompeu Fabra \\ 
}

\begin{document}
\maketitle

\begin{abstract}
Generalized planning aims at computing an algorithm-like structure ({\em generalized plan}) that solves a set of multiple planning instances. In this paper we define {\em negative examples} for generalized planning as planning instances that must not be solved by a generalized plan. With this regard the paper extends the notion of {\em validation} of a generalized plan as the problem of verifying that a given generalized plan solves the set of input {\em positives} instances while it fails to solve a given input set of {\em negative} examples. This notion of plan validation allows us to define quantitative metrics to asses the generalization capacity of generalized plans. The paper also shows how to incorporate this new notion of plan validation into a compilation for plan synthesis that takes both positive and negative instances as input. Experiments show that incorporating negative examples can accelerate plan synthesis in several domains and leverage quantitative metrics to evaluate the generalization capacity of the synthesized plans.
 \end{abstract}

\section{Introduction}
{\em Generalized planning} studies the computation of plans that can solve a family of planning instances that share a common structure~\cite{hu2011generalized,srivastava2011new,jimenez2019review}. Since {\em generalized planning} is computationally expensive, a common approach is to synthesize a generalized plan starting from a set of small instances and validate it on larger instances. This approach is related to the principle of Machine Learning (ML), in which a model is trained on a {\em training set} and validated on a {\em test set}~\cite{Mitchell:ML:1997}.

Traditionally, generalized planning has only focused on solvable instances, both for plan synthesis and for validation~\cite{Winner03distill:learning,Geffner:FSM:AAAI10,Levesque:GPlanning:IJCAI11,Zilberstein:Gplanning:icaps11,Giacomo:FSM:ICAPS13,segovia2018computing,segovia2019computing}. However, many computational problems in AI benefit from {\em negative} examples, e.g. SAT approaches that exploit clause learning~\cite{angluin1992learning}, grammar induction~\cite{parekh2000grammar}, program synthesis~\cite{alur2018search} and model learning~\cite{camacho2019models}. If used appropriately, negative examples can help reduce the solution space and accelerate the search for a solution.

\begin{figure} 
\begin{subfigure}[a]{0.43\columnwidth}
\begin{tiny}     
\vspace{-3cm}
\begin{tikzpicture}[scale=.5]
        \begin{scope}
          \draw (0, 0) grid (2, 1);
          \node[fill=black,minimum size=.5cm] at (0.5, 0.5) {};
          \node[anchor=center] at (1.5, 0.5) {R};
        \end{scope}
      \end{tikzpicture}          

\vspace{.7cm}
\begin{tikzpicture}[scale=.5]
        \begin{scope}
          \draw (0, 0) grid (6, 1);
          \node[fill=black,minimum size=.5cm] at (0.5, 0.5) {};
          \node[anchor=center] at (1.5, 0.5) {};
          \node[fill=black,minimum size=.5cm] at (2.5, 0.5) {};
          \node[anchor=center] at (3.5, 0.5) {};
           \node[fill=black,minimum size=.5cm] at (4.5, 0.5) {};
          \node[anchor=center] at (5.5, 0.5) {R};
        \end{scope}
      \end{tikzpicture}

\vspace{.7cm}
\begin{tikzpicture}[scale=.5]
        \begin{scope}
          \draw (0, 0) grid (1, 1);
           \node[minimum size=.5cm] at (0.5, 0.5) {};          
          \node[anchor=center] at (0.5, 0.5) {R};
        \end{scope}
      \end{tikzpicture}
\vspace{2cm}      
\end{tiny}
\centerline{a) \scriptsize Three goal configurations.}
\end{subfigure}
\begin{subfigure}[b]{0.5\columnwidth}
\begin{tiny}
$\Pi$:
\begin{tikzpicture}[->,>=stealth',shorten >=1pt,auto,node distance=1.4cm,semithick]
	  \node[state] (A)              {$q_0$};
	  \node[state] (B) [right of=A] {$q_1$};
	  \node[state] (C) [right of=B] {$q_2$};
	  \node[state] (D) [right of=C] {$\overline{q_3}$};
	  \path
          (A) edge [align=center] node {$\mathsf{paint}(X)$} (B)
          (B) edge [align=center] node {$\mathsf{inc}(X)$} (C)
          (C) edge [align=center] node {$\mathsf{inc}(X)$} (D);
    \end{tikzpicture}

\vspace{.7cm}
$\Pi^{*}$:
\begin{tikzpicture}[->,>=stealth',shorten >=1pt,auto,node distance=1.4cm,semithick]
	  \node[state] (A)              {$\overline{q_0}$};
	  \node[state] (B) [right of=A] {$q_1$};
	  \node[state] (C) [right of=B] {$q_2$};
	  \path
          (A) edge [align=center] node {$\mathsf{paint}(X)$} (B)
          (B) edge [align=center] node {$\mathsf{inc}(X)$} (C)
          (C) edge [align=center, bend left] node {$\mathsf{inc}(X)$} (A);                                        
    \end{tikzpicture}

\vspace{.7cm}
$\Pi^{+}$:
\begin{tikzpicture}[->,>=stealth',shorten >=1pt,auto,node distance=1.4cm,semithick]
	  \node[state] (A)              {$q_0$};
	  \node[state] (B) [right of=A] {$q_1$};
	  \node[state] (C) [right of=B] {$q_2$};
	  \node[state] (D) [right of=C] {$\overline{q_3}$};          
	  \path
          (A) edge [align=center] node {$\mathsf{paint}(X)$} (B)
          (B) edge [align=center] node {$\mathsf{inc}(X)$} (C)
          (C) edge [align=center] node {$\mathsf{inc}(X)$} (D)
          (D) edge [align=center, bend left] node {$\mathsf{paint}(X)$} (B);                    
    \end{tikzpicture}
\end{tiny}
\centerline{b) \scriptsize Algorithm-like plans.}
\end{subfigure}                                       
\caption{\small a) Robot in $2\times 1$, $6\times 1$ and $1\times 1$ corridors; b) Three candidate generalized plans.}
  \label{fig:basicprograms}
\end{figure}
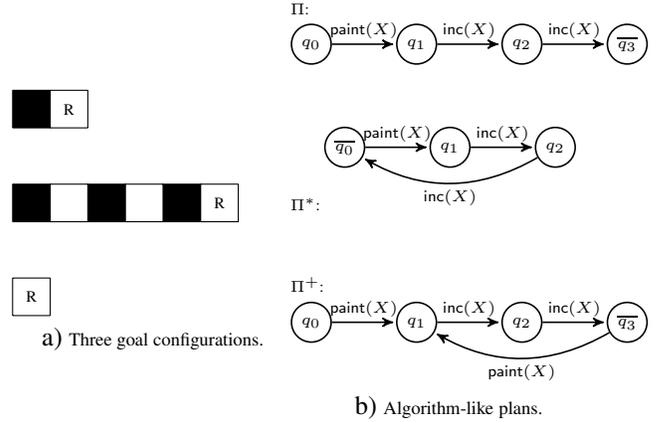

In this paper we introduce \textit{negative examples} for generalized planning as input planning instances that should not be solved by a generalized plan. An intuitive way to come up with negative examples for solutions that generalize is to first synthesize a solution with exclusively positive examples, and identify cases for which the solution did not generalize as desired, somewhat akin to {\em clause learning} in satisfiability problems~\cite{biere2009conflict}. Imagine that we aim to synthesize the generalized plan {\small $(\mathsf{paint}(X),\mathsf{inc}(X),\mathsf{inc}(X))^+$} that makes a robot paint every odd cell black in a $N\times 1$ corridor, starting from the leftmost cell (we use Kleene notation to represent regular expressions, and $Z^+$ indicates the repetition of $Z$ at least once). Action {\small $\mathsf{paint}(X)$} paints the current cell black while {\small $\mathsf{inc}(X)$} increments the robot's $X$ coordinate. The {\em positive example} of a $2\times 1$ corridor (whose goal configuration is illustrated at Figure~\ref{fig:basicprograms}a) is solvable by all three automata plans in Figure~\ref{fig:basicprograms}b) (acceptor states are marked with overlines e.g., $\overline{q}$). These automata plans, namely $\Pi$, $\Pi^{*}$ and $\Pi^{+}$, compactly represent the three sets of sequential plans {\small$(\mathsf{paint}(X),\mathsf{inc}(X),\mathsf{inc}(X))$}, {\small $(\mathsf{paint}(X),\mathsf{inc}(X),\mathsf{inc}(X))^*$} and {\small$(\mathsf{paint}(X),\mathsf{inc}(X),\mathsf{inc}(X))^+$}. Hence the single positive example of a $2\times 1$ corridor is not enough to discriminate among these three generalized plans. Adding a second {\em positive example}, the $6\times 1$ corridor in Figure~\ref{fig:basicprograms}a), discards plan $\Pi$. Adding a third $1\times 1$ {\em negative example}, where the initial and goal robot cell are the same and no cell is required to be painted, discards $\Pi^{*}$ preventing {\em over-generalization} because $\Pi^{*}$ solves this negative example. 

The problem of deriving generalized plans has been a longstanding open challenge. Compared to previous work on generalized planning, the contributions of this paper are:
\begin{enumerate}
    \item \textbf{Negative examples} to more precisely specify the semantics of an aimed generalized plan.
    \item A new approach for the synthesis of plans that can generalize from \textbf{smaller input examples} thanks to negative examples.
    \item The definition of quantitative  \textbf{evaluation metrics} to assess the generalization capacity of generalized plans.        
\end{enumerate}

The paper is organized as follows. We start with some background notation of classical planning and generalized planning (GP). Then we formalize the concept of a {\em negative example} for generalized planning. We continue with a description of a generalized planning problem with positive and negative examples that can be compiled to classical planning. We show proofs of soundness and completeness for the two main tasks in GP that are synthesis and validation. We continue with the experiments where we compare the impact of negative examples, and finally we conclude with a discussion on the presented work.

\section{Background}
\label{sec:background}
This section formalizes the planning models used in this work as well as {\em planning programs}~\cite{segovia2019computing}, an algorithm-like representation for plans that can generalize.

\subsection{Classical planning with conditional effects}
We use $F$ to denote the set of {\em fluents} (propositional variables) describing a state. A literal $l$ is a valuation of a fluent $f\in F$, i.e.~$l=f$ or $l=\neg f$. A set of literals $L$ on $F$ represents a partial assignment of values to fluents (WLOG we assume that $L$ does not assign conflicting values to any fluent). Given $L$, $\neg L=\{\neg l:l\in L\}$ is the complement of $L$. Finally, we use $\mathcal{L}(F)$ to denote the set of all literal sets on $F$, i.e.~all partial assignments of values to fluents. A {\em state} $s$ is a set of literals such that $|s|=|F|$, i.e.~a total assignment of values to fluents. The number of states is then $2^{|F|}$. 

A {\em classical planning frame} is a tuple $\Phi=\tup{F,A}$, where $F$ is a set of fluents and $A$ is a set of actions with {\em conditional effects}. Conditional effects can compactly define actions whose precise effects depend on the state where the action is executed.  Each action $a\in A$ has a set of literals $\pre(a)$, called the {\em precondition}, and a set of {\em conditional effects}, $\cond(a)$. Each conditional effect $C\rhd E\in\cond(a)$ is composed of a set of literals $C$ (the condition) and $E$ (the effect). Action $a$ is {\em applicable} in state $s$ if and only if $\pre(a)\subseteq s$, and the resulting set of {\em triggered effects} is
\[
\eff(s,a)=\bigcup_{C\rhd E\in\cond(a),C\subseteq s} E,
\]
i.e.~effects whose conditions hold in $s$. The result of applying $a$ in $s$ is the {\em successor state} $\theta(s,a)=(s\setminus \neg\eff(s,a))\cup\eff(s,a)$.

A {\em classical planning problem} with conditional effects is a tuple $P=\tup{F,A,I,G}$, where $F$ is a set of fluents and $A$ is a set of actions with {\em conditional effects} as defined for a {\em planning frame}, $I$ is an initial state and $G$ is a goal condition, i.e.~a set of literals to achieve.

A {\em solution} for a classical planning problem $P$ can be specified using different representation formalisms, e.g. a sequence of actions, a partially ordered plan, a policy, etc. Here we define a {\em sequential plan} for $P$ as an action sequence $\pi=\tup{a_1, \ldots, a_n}$ whose execution induces a state sequence $\tup{s_0, s_1, \ldots, s_n}$ such that $s_0=I$ and, for each $i$ such that $1\leq i\leq n$, $a_i$ is applicable in $s_{i-1}$ and generates the successor state $s_i=\theta(s_{i-1},a_i)$. The plan $\pi$ {\em solves} $P$ if and only if $G\subseteq s_n$, i.e.~if the goal condition is satisfied following the execution of $\pi$ in $I$.



\subsection{Planning programs}
Given a planning frame $\Phi=\tup{F,A}$, a {\em planning program}~\cite{segovia2019computing} is a sequence of instructions $\Pi=\tup{w_0,\ldots,w_n}$. Each instruction $w_i$, {\small $0\leq i\leq n$}, is associated with a {\em program line} $i$ and is drawn from the set of instructions $\mathcal{I}=A\cup\mathcal{I}_{go}\cup\{\mathsf{end}\}$, where $\mathcal{I}_{go}=\{\,\mathsf{go}(i',!f):\,0\leq i'\leq n,f\in F\}$ is the set of {\em goto instructions}. In other words, each instruction is either a planning action $a\in A$, a goto instruction $\mathsf{go}(i',!f)$ or a termination instruction $\mathsf{end}$. 

The execution model for a planning program $\Pi$ is a {\em program state} $(s,i)$, i.e.~a pair of a planning state $s\subseteq\mathcal{L}(F)$ (with $|s|=|F|$), and a program counter $0\leq i\leq n$. Given a program state $(s,i)$, the execution of instruction $w_i$ on line $i$ is defined as follows:
\begin{itemize}
\item If $w_i\in A$, the new program state is $(s',i+1)$, where $s'=\theta(s,w)$ is the {\em successor} for planning state $s$ and action $w$.
\item If $w_i=\mathsf{go}(i',!f)$, the program state becomes $(s,i+1)$ if $f\in s$, and $(s,i')$ otherwise. Conditions in Goto instructions can represent arbitrary formulae since $f$ can be a {\em derived fluent}~\cite{lotinac2016automatic}.
\item If $w_i=\mathsf{end}$, execution terminates.
\end{itemize}

To execute a planning program $\Pi$ on a planning problem $P=\tup{F,A,I,G}$, the initial program state is set to $(I,0)$, i.e.~the initial state of $P$ and program line $0$. A program $\Pi=\tup{w_0,\ldots,w_n}$ {\em solves} a planning problem $P=\tup{F,A,I,G}$ iff the execution terminates in a program state $(s,i)$ that satisfies the goal conditions, i.e.~$w_i=\mathsf{end}$ and $G\subseteq s$.

\citeauthor{segovia2019computing}~(\citeyear{segovia2019computing}) contains a detailed analysis of the failure conditions on planning programs, which we summarize here as follows:

\begin{corollary}[Planning Program Failure]
If a planning program $\Pi$ {\em fails} to solve a planning problem $P$, the only possible sources of failure are:
\begin{enumerate}
\item {\bf Incomplete Program}. Execution terminates in program state $(s,i)$ but the goal condition does not hold, i.e.~($w_i=\mathsf{end})\wedge (G\not\subseteq s)$.
\item {\bf Inapplicable Action}. Executing an action $w_i\in A$ in program state $(s,i)$ fails because its precondition does not hold, i.e.~$\pre(w)\not\subseteq s$.
\item {\bf Infinite Loop}. The program execution enters into an infinite loop that never reaches an $end$ instruction.
\end{enumerate}
\end{corollary}



\subsection{Generalized planning}
We define a {\em generalized planning problem} as a finite set of classical planning problems $\mathcal{P}=\{P_1,\ldots,P_T\}$ that are defined on the same planning frame $\Phi$. Therefore, $P_1=\tup{F,A,I_1,G_1},\ldots,P_T=\tup{F,A,I_T,G_T}$ share the same fluents and actions but differ in the initial state and goals. 

A {\em planning program} $\Pi$ solves a given generalized planning problem $\mathcal{P}$ iff $\Pi$ solves every planning problem $P_t\in \mathcal{P}$, {\small $1\leq t\leq T$}.

\citeauthor{segovia2019computing}~(\citeyear{segovia2019computing}) showed that a program $\Pi$ that solves a generalized planning task $\mathcal{P}$ can be synthesized by defining a new classical planning problem $P_n=\tup{F_n,A_n,I_n,G_n}$, where $n$ is a bound on the number of program lines. A solution plan $\pi$ for $P_n$ programs instructions on the available empty lines (building the n-line program $\Pi$), and validates $\Pi$ on each input problem $P_t\in\mathcal{P}$. 

The {\bf fluent set} is defined as $F_n=F\cup F_{pc}\cup F_{ins}\cup F_{test}\cup\{\mathsf{done}\}$, where:
\begin{itemize}
\item $F_{pc} = \{\mathsf{pc}_i : 0 \le i \le n\}$ models the {\em program counter},
\item $F_{ins} = \{\mathsf{ins}_{i,w} : 0 \le i \le n, w \in \mathcal{I}\cup\{\mathsf{nil}\}\}$ models the program lines ($\mathsf{nil}$ indicates an empty line),
\item $F_{test} = \{\mathsf{test}_t : 1\leq t\leq T\}$ indicates the classical planning problem $P_t\in \mathcal{P}$.
\end{itemize}

Each instruction $w\in\mathcal{I}$ is modeled as a planning action, with one copy $\mathsf{end}_t$ of the termination instruction per input problem $P_t$. Preconditions for the goto and end instructions are defined as $\pre(\mathsf{go}(i',!f))=\emptyset$ and $\pre(\mathsf{end}_t)=G_t\cup\{\mathsf{test}_t\}$. The authors define two {\bf actions} for each instruction $w$ and line $i$: a {\em programming action} $P(w_i)$ for programming $w$ on line $i$, and an {\em execution action} $E(w_i)$ that uses the previous execution model to execute $w$ on line $i$:
\begin{align*}
\pre(P(w_i))&=\pre(w)\cup\{\mathsf{pc}_i,\mathsf{ins}_{i,\mathsf{nil}}\},\\
\cond(P(w_i))&=\{\emptyset\rhd\{\neg\mathsf{ins}_{i,\mathsf{nil}},\mathsf{ins}_{i,w}\}\},\\
\pre(E(w_i))&=\pre(w)\cup\{\mathsf{pc}_i,\mathsf{ins}_{i,w}\}.
\end{align*}
The effect of $E(w_i)$ depends on the type of instruction:
\begin{align*}
\cond(E(w_i))&=\cond(w)\cup\{\emptyset\rhd\{\neg\mathsf{pc}_i,\mathsf{pc}_{i+1}\} \}, w\in A,\\
\cond(E(w_i))&=\{\emptyset\rhd\{\neg\mathsf{pc}_i\}\} \cup \{\{f\}\rhd\{\mathsf{pc}_{i+1}\}\}  \\
&\hspace*{0.4cm}\cup \{\{\neg f\}\rhd\{\mathsf{pc}_{i'}\}\}, \hspace*{0.5cm} w=\;\mathsf{go}(i',!f),&\\
\cond(E(w_i))&=\mathsf{reset}_{t+1}, \hspace*{2.15cm} w=\;\mathsf{end}_t,t<T,\\
\cond(E(w_i))&=\{\emptyset\rhd\{\mathsf{done}\}\}, \hspace*{1.32cm} w=\;\mathsf{end}_T.
\end{align*}
The conditional effect $\mathsf{reset}_{t+1}$ resets the program state to $(I_{t+1},0)$, preparing execution on the next problem $P_{t+1}$.

The {\bf initial state} is $I_n=I_1\cup\{\mathsf{pc}_0\}\cup\{\mathsf{ins}_{i,\mathsf{nil}}:0\leq i\leq n\}$ indicating that, initially, the program lines are empty and the program counter is at the first line. The {\bf goal} is $G_n=\{\mathsf{done}\}$ and can only be achieved after solving sequentially all the instances in the generalized planning problem.

\begin{figure}
    \centering
    \begin{tikzpicture}
    \draw[draw=black,step=0.5cm] (0.0,0.0) grid (3.0,0.5);
    \draw[draw=black,step=0.5cm] (4.0,0.0) grid (7.0,0.5);
    \draw[draw=black] (4.0,0.0) -- (4.0,0.5);
    \draw[draw=black,step=0.5cm] (0.0,1.0) grid (3.0,1.5);
    \draw[draw=black,step=0.5cm] (4.0,1.0) grid (7.0,1.5);
    \draw[draw=black] (4.0,1.0) -- (4.0,1.5);
    \draw[draw=black] (0.0,1.0) -- (3.0,1.0);
    \draw[draw=black] (4.0,1.0) -- (7.0,1.0);
    \draw[->,black] (3.1,0.25) -- (3.9,0.25);
    \draw[->,black] (3.1,1.25) -- (3.9,1.25);
    \node at (1.5,2.0) {\textbf{Initial State}};
    \node at (5.5,2.0) {\textbf{Goal State}};
    \node at (0.25,0.25) {1};
    \node at (0.75,0.25) {2};
    \node at (1.25,0.25) {3};
    \node at (1.75,0.25) {4};
    \node at (2.25,0.25) {5};
    \node at (2.75,0.25) {6};
    \node at (4.25,0.25) {6};
    \node at (4.75,0.25) {5};
    \node at (5.25,0.25) {2};
    \node at (5.75,0.25) {3};
    \node at (6.25,0.25) {4};
    \node at (6.75,0.25) {1};
    \node at (0.25,1.25) {1};
    \node at (0.75,1.25) {2};
    \node at (1.25,1.25) {3};
    \node at (1.75,1.25) {4};
    \node at (2.25,1.25) {5};
    \node at (2.75,1.25) {6};
    \node at (4.25,1.25) {6};
    \node at (4.75,1.25) {5};
    \node at (5.25,1.25) {4};
    \node at (5.75,1.25) {3};
    \node at (6.25,1.25) {2};
    \node at (6.75,1.25) {1};
    \end{tikzpicture}
    \caption{\small {\em Positive example} (upper row) and {\em negative example} (lower row) for the generalized planning task of reversing a list.}
    \label{fig:pn-gp-example}
\end{figure}

\section{Negative examples in generalized planning}
\label{sec:gp-pn}

This section extends the previous generalized planning formalism to include {\em negative examples} for the validation and synthesis of programs.

\subsection{Negative examples as classical planning problems}

Negative examples are additional solution constraints to more precisely specify the semantics of an aimed generalized plan and prevent undesired generalizations.

\begin{definition}[Negative examples in generalized planning] Given a generalized planning problem $\mathcal{P}$, a negative example is a classical planning instance $P^-=\tup{F,A,I^-,G^-}$ that must not be solved by solutions to $\mathcal{P}$.
\end{definition}

Figure~\ref{fig:pn-gp-example} shows an input/output pair $(1,2,3,4,5,6)/(6,5,4,3,2,1)$ that represents a positive example for computing a generalized plan that reverse lists of any size. This example can be encoded as a classical planing problem, where the set of fluents are the state variables necessary for encoding a list of arbitrary size plus two pointers over the list nodes. The initial state encodes, using these fluents, the particular list $(1,2,3,4,5,6)$. The goal condition encodes the target list $(6,5,4,3,2,1)$. Finally, actions encode the swapping of the content of two pointers as well as actions for moving the pointers along the list. In this regard, the input/output example $(1,2,3,4,5,6)/(6,5,4,3,2,1)$ is a {\em positive example} while $(1,2,3,4,5,6)/(6,5,2,3,4,1)$ or $(4,3,2,1)/(2,3,4,1)$ are {\em negative examples} for the generalized planning task of reversing lists.

In this work both {\em positive} and {\em negative} examples are classical planning problems $P_1=\tup{F,A,I_1,G_1},\ldots,P_T=\tup{F,A,I_T,G_T}$ defined on the same fluent set $F$ and action set $A$. Each problem $P_t\in \mathcal{P}$, {\small $1\leq t\leq T$} encodes an {\em input} specification in its initial state $I_t$ while $G_t$ encodes the specification of its associated {\em output}. Although the examples share actions, each action in $A$ can have different interpretations in different states due to {\em conditional effects}. For instance, back to the example of Figure~\ref{fig:basicprograms}, we can encode individual planning tasks with different corridor sizes (the set of fluents $F$ can include fluents of type $\mathsf{max}(N)$ that encode different corridor boundaries~\cite{segovia2019computing}).

Negative examples should not be confused with {\em unsolvable planning instances}. The goals of negative examples are reachable from the given initial state (see Figure~\ref{fig:pn-gp}). For instance the goals of the negative example $(1,2,3,4,5,6)/(6,5,2,3,4,1)$ shown in Figure~\ref{fig:pn-gp-example} can be reached from the associated initial state by applying the corresponding actions to swap the content of pointers and moving appropriately those pointers. On the other hand $(4,3,2,1)/(1,1,1,1)$ would represent an UNSAT classical planning instance, because $(1,1,1,1)$ is not reachable starting from $(4,3,2,1)$ and provided the mentioned actions for reversing lists.  

\begin{figure}
\centering
\begin{tikzpicture}
\draw[draw=black] (0.0,0.0) rectangle (8.0,3.0);
\draw[draw=black] (0.0,2.0) -- (4.0,2.0);
\draw[draw=black] (0.0,1.0) -- (4.0,1.0);
\draw[draw=black] (4.0,0.0) -- (4.0,3.0);
\draw[draw=black] (1.9,0.0) -- (1.9,2.0);
\node at (6.0,1.5) {\Large UNSAT};
\node at (2.0,2.5) {\Large SAT};
\node at (1,1.5) {\small Positives};
\node at (1,0.5) {\scriptsize $P_{1^+},\ldots,P_{T{^+}}$};
\node at (3,1.5) {\small Negatives};
\node at (3,0.5) {\scriptsize $P_{T^+ + 1},\ldots,P_{T}$};
\end{tikzpicture}
\caption{\small Taxonomy of instances in {\em generalized planning}.}
\label{fig:pn-gp}
\end{figure}



\subsection{Program validation with negative examples}
\label{ssec:valPosAndNeg}

In generalized planning the process of {\em plan validation} is implicitly required as part of {\em plan synthesis}, since computing a solution plan requires us to validate it on all the given input instances. Next, we extend the notion of validation to consider also negative examples.

\begin{definition}[Program Validation with {\em Positive} and {\em Negative} examples]
Given a program $\Pi$ and a set of classical planning problems $\mathcal{P}=\{P_1,\ldots,P_T\}$ labeled either as {\em positive} or {\em negative}, {\em validation} is the task of verifying whether $\Pi$ solves each $P\in\mathcal{P}$ labeled as {\em positive}, while it fails to solve each $P\in\mathcal{P}$ that is labeled as {\em negative}.
\end{definition}

Validating a sequential plan on a classical planing problem is straightforward because either a {\em validation} proof, or a {\em failure} proof, is obtained by executing the plan starting from the initial state of the planning problems~\cite{howey2004val}. Validating a program on a classical planning problem is no longer straightforward because it requires a mechanism for detecting {\em infinite loops} (checking failure conditions 1. and 2. is however straightforward).

The execution of plans with control flow on a given planning problem is compilable into classical planning. Examples are compilations for {\em {\sc GOLOG} procedures}~\cite{baier2007exploiting}, {\em Finite State Controllers}~\cite{Geffner:FSM:AAAI10,segovia2018computing} or {\em planning programs}~\cite{javi-Gplanning-ICAPS16,segovia2019computing}. These compilations encode the {\em cross product} of the given planning problem and the automata corresponding to the plan to execute. The plan is {\em valid} iff the compiled problem is solvable. If a classical planner proves the compiled problem is unsolvable, then the plan is {\em invalid} because its execution necessarily failed.

Besides current planners do not excel at proving that a given problem is {\em unsolvable}~\cite{eriksson2017unsolvability}, none of the cited compilations can identify the precise source of a failed plan execution. Next, we show that the classical planning compilation for the synthesis of {\em planning programs}~\cite{segovia2019computing} can be updated with a mechanism for detecting infinite loops, that is taken from an approach for the computation of infinite plans~\cite{nir:infiniteplans:IJCAI11}. This updated compilation can identify the three possible sources of execution failure (namely {\bf incomplete programs}, {\bf inapplicable actions} and {\bf infinite loops}) of a program in a given classical planning problem. What is more, the compilation can be further updated for solving {\em generalized planning problems with positive and negative examples}.

\subsection{A compilation for program validation}
\label{ssec:comp2val}

Given a generalized planning task $\mathcal{P}=\{P_1,\ldots,P_T\}$ and a program $\Pi$, program validation is compilable into a planning instance $P_n'=\tup{F_n',A_n',I_n',G_n}$, that extends $P_n$ from the {\em background} Section~\ref{sec:background}. The extended fluent set is $F_n'=F_n\cup F_{neg}\cup F_{copy}$, where
\begin{itemize}
\item $F_{neg} = \{\mathsf{checked, holds, stored, acted, loop}\}$ contains flags for identifying the source of execution failures,
\item $F_{copy} = \{\mathsf{copy}_f, \mathsf{correct}_f : f \in F \cup F_{pc}\}$ contains the fluents used to store a copy of the program state with the aim of identifying infinite loops.
\end{itemize}

Unlike $A_n$, the action set $A_n'$ does not contain {\em programming action} (these actions are only necessary for program synthesis but not for program validation). However, $A_n'$ contains a new type of action called {\em check action} that verifies whether the precondition of an instruction holds. For an instruction $w$ and line $i$, the check action $C(w_i)$ is defined as
\begin{align*}
  \pre(C( w_i ))=& \{\mathsf{pc}_i,\mathsf{ins}_{i,w},\neg\mathsf{checked},\neg\mathsf{loop}\}, \\
  \cond(C( w_i ))=& \{\emptyset\rhd\{ \mathsf{checked} \}\}\cup \{\pre(w)\rhd\{\mathsf{holds}\}\}.
\end{align*}
After applying $C(w_i)$, execution fails if $\mathsf{holds}$ is false, either because the goal condition $G_t$ is not satisfied when we apply a termination instruction $\mathsf{end}_t$, or because the precondition $\pre(w)$ of the action $w\in A$ does not hold (which corresponds precisely to failure conditions 1. and 2. above). A similar mechanism has been previously developed for computing {\em explanations/excuses} of when a plan cannot be found~\cite{gobelbecker2010coming}.

Each execution action $E(w_i)$ is defined as before, but we add precondition $\{\mathsf{checked},\mathsf{holds}\}$ and the conditional effect $\emptyset\rhd\{\neg\mathsf{checked},\neg\mathsf{holds},\mathsf{acted}\}$. As a result, $C(w_i)$ has to be applied before $E(w_i)$, and $E(w_i)$ is only applicable if execution does not fail (i.e.~if $\mathsf{holds}$ is true).

To identify {\em infinite loops} $A_n'$ is extended with three new actions:
\begin{itemize}
\item $\mathsf{store}$, which stores a copy of the current program state.
\begin{align*}
  \pre(\mathsf{store})&= \{\neg\mathsf{checked},\neg\mathsf{stored},\mathsf{acted}\}, \\
  \cond(\mathsf{store})&=\{\emptyset\rhd\{\mathsf{stored},\neg\mathsf{acted}\}\}\\
				&\;\cup\{ \{f\}\rhd \{\mathsf{copy}_f\} : \forall f \in F \cup F_{pc} \}.
\end{align*}
This action can be applied only once, after an action execution $E(w_i)$ and before checking an action $C(w_i)$. It simply uses conditional effects to store a copy of the program state $(s,i)$ in the fluents of type $\mathsf{copy}_f$.

\item $\mathsf{compare}$, which compares the current program state $(s,i)$ with the stored copy.
\begin{align*}
  \pre&(\mathsf{compare})= \{\neg\mathsf{checked},\mathsf{stored},\mathsf{acted},\neg\mathsf{loop}\},\\
  \cond&(\mathsf{compare})=\{\emptyset\rhd\{\neg\mathsf{stored},\neg\mathsf{acted},\mathsf{loop}\}\}\\
					  \cup &\{ \{f, \mathsf{copy}_f \} \rhd \{ \mathsf{correct}_f \} : f \in F \cup F_{pc} \} \\
					   \cup&\{ \{\neg f, \neg\mathsf{copy}_f \} \rhd \{ \mathsf{correct}_f \} : f \in F \cup F_{pc} \}.
\end{align*}
The result of the comparison is in the fluents of type $\mathsf{correct}_f$. Note that $\mathsf{acted}$ is not true after applying $\mathsf{store}$ so, to satisfy the precondition of $\mathsf{compare}$, we have to apply an execution action first (otherwise the current program state would trivially equal the stored copy). For a fluent $f$ to be correct, either it is true in both the current program state and the stored copy, or it is false in both.

\item $\mathsf{process}$, which processes the outcome of the comparison.
\begin{align*}
  \pre&(\mathsf{process})= \{\mathsf{loop}\}\cup\{\mathsf{correct}_f:f \in F \cup F_{pc}\},\\
  \cond&(\mathsf{process})=\{\emptyset\rhd\{\neg\mathsf{loop},\mathsf{checked}\}\}.
\end{align*}
This action can only be applied if all fluents in $F \cup F_{pc}$ are correct, adds fluent $\mathsf{checked}$, and resets other auxiliary fluents to false. The purpose of adding $\mathsf{checked}$ is to match the state of other failure conditions ($\mathsf{checked}$ is true and $\mathsf{holds}$ is false).
\end{itemize}

Finally, $A_n'$ contain also actions $\mathsf{skip}_t$, {\small $1\leq t\leq T$}, that terminate program execution as a result of a failure condition. These actions are applicable once a failure condition is detected, of either type ($\mathsf{checked}$ is true and $\mathsf{holds}$ is false). 
\begin{align*}
  \pre(\mathsf{skip}_t)&= \{\mathsf{test}_t,\mathsf{checked},\neg\mathsf{holds}\}, \\
  \cond(\mathsf{skip}_t)&= \cond(\mathsf{end}_t)\\
  &\cup\{\emptyset\rhd\{\neg\mathsf{checked},\neg\mathsf{stored}\}\}\\
  &\cup \{\emptyset\rhd\{\neg\mathsf{copy}_f,\neg\mathsf{correct}_f: f \in F \cup F_{pc} \}\}.
\end{align*}

Note that the action applied immediately before $\mathsf{skip}_t$ identifies the source of the execution failure of the program $\Pi$ on $P_t$. This action is either:
\begin{enumerate}
\item $C(\mathsf{end}_{t,i})$, identifying an {\bf incomplete program}.
\item $C(w_i)$ such that $w\in A$, which proves that $w\in A$ is an {\bf inapplicable action}.
\item $\mathsf{process}$, identifying that the execution of the program entered an {\bf infinite loop}.
\end{enumerate}

The precondition $\neg\mathsf{stored}$ is added to all check actions $C(\mathsf{end}_{t,i})$, to avoid saving a stored copy of the program state from one input instance to the next. The goal condition is the same as before and in the initial state $I_n'$ the instructions of the program $\Pi$ are already programmed in the initial state:
\[I_n'=I_1\cup\{\mathsf{pc}_0\}\cup\{\mathsf{ins}_{i,w}:0\leq i\leq n \wedge w_i\in\Pi\}.\]

\subsection{Program synthesis with positive and negative examples}
\label{ssec:synthesisnegex}

We define a  {\em generalized planning problem with positive and negative examples} as a finite set of classical planning instances $\mathcal{P} = \{P_1,\ldots,P_{T^+},\ldots,P_T\}$ that belong to the same planning frame. In this set there are $T^+$ positive and $T^-$ negative instances such that $T = T^+ + T^-$ (see Figure~\ref{fig:pn-gp}). We assume that at least one positive instance is necessary ($T^+>0$) because otherwise, the one-instruction program $0. end$, covers any negative instance whose goals are not satisfied in the initial state.

To synthesize programs for generalized planning with positive and negative examples we extend the compilation $P_n'$ with {\em programming instructions}. The output of the final extension of the compilation is a new planning instance $P_n'' = \tup{F_n'',A_n'',I_n'',G_n}$:
\begin{itemize}
\item The fluent set $F_n''$ is identical to that of the compilation $P_n'$, except that $F_n''$ now includes a fluent $\mathsf{negex}$, which is used to constrain the application of actions $E(\mathsf{end}_{t,i})$ and $\mathsf{skip}_t$, respectively. By adding a precondition $\neg\mathsf{negex}$ to $E(\mathsf{end}_{t,i})$ and a precondition $\mathsf{negex}$ to $\mathsf{skip}_t$, we require program execution to succeed for positive examples, and to fail for negative examples.  

\item The action set $A_n''$ is identical to $A_n'$ except that we reintroduce {\em programming actions} $P(w_i)$ in the action set $A_n''$ and we add a precondition $\neg\mathsf{negex}$ to $E(\mathsf{end}_{t,i})$ and a precondition $\mathsf{negex}$ to $\mathsf{skip}_t$ to require program execution to succeed for positive examples, and to fail for negative examples. Moreover, precondition $\mathsf{negex}$ is added to all actions related to infinite loop detection (e.g.  $\mathsf{store},\mathsf{compare}$ and $\mathsf{process}$).
\item All program lines are now empty in $I_n''$ (so they can be programmed) and the goal condition is still $G_n=\{\mathsf{done}\}$, like in the original compilation. 
 \end{itemize}

\begin{theorem}[Soundness]
Any plan $\pi$ that solves the planning instance $P_n''$ induces a planning program $\Pi$ that solves the corresponding generalized planning task with positive and negative examples $\mathcal{P} = \{P_1,\ldots,P_{T^+},\ldots,P_T\}$.
\end{theorem}

\begin{small}
\begin{proof}
To solve the planning instance $P_n''$, a solution $\pi$ has to use {\em programming actions} $P(w_i)$ to program the instructions on empty program lines, effectively inducing a planning program $\Pi$. Once an instruction $w$ is programmed on line $i$, it cannot be altered and can only be executed using the given execution model (that is, using a {\em check action} $C(w_i)$ to test its precondition, followed by an {\em execution action} $E(w_i)$ to apply its effects). To achieve the goal $G_n=\{\mathsf{done}\}$, $\pi$ has to simulate the execution of $\Pi$ on each input instance $P_t$, {\small $1\leq t\leq T$}, terminating with either $E(\mathsf{end}_{t,i})$ or $\mathsf{skip}_t$, which are the only two actions that allow us to move on to the next input instance (if $t<T$) or add fluent $\mathsf{done}$ (if $t=T$). Because of the preconditions of $E(\mathsf{end}_{t,i})$ and $\mathsf{skip}_t$, termination has to end with $E(\mathsf{end}_{t,i})$ if $P_t$ is a positive example, and with $\mathsf{skip}_t$ if $P_t$ is negative proving that $\Pi$ solves each positive example and fails to solve each negative example.
\end{proof}
\end{small}

\begin{theorem}[Completeness]
If there exists a program $\Pi$ within $n$ program lines that solves $\mathcal{P} = \{P_1,\ldots,P_{T^+},\ldots,P_T\}$ then there exists a classical plan $\pi$ that solves $P_n''$.
\end{theorem}

\begin{small}
\begin{proof}
We can build a prefix of plan $\pi$ using programming actions that insert the instructions of $\Pi$ on each program line. Now, we determine the remaining actions of $\pi$ building a postfix that simulates the execution of $\Pi$ on each input instance $P_t\in \mathcal{P}$. Since $\Pi$ solves each positive example and fails to solve each negative example, it means that there exists an action sequence that simulates the execution of $\Pi$ on $P_t$ and ends with action $E(\mathsf{end}_{t,i})$ (if $P_t$ is a positive example) and with $\mathsf{skip}_t$ (if $P_t$ is negative). 
\end{proof}
\end{small}

\section{Experiments}
This section reports the empirical performance of our approach for the {\em synthesis} and {\em evaluation} of programs for generalized planning\footnote{The source code, benchmarks and scripts are in the Automated Programming Framework \cite{automated-programming-framework} such that any experimental data in the paper can be reproduced.}. All experiments are run on an {\em Intel Core i5 2.90GHz x 4} with a memory limit of 4GB and 600 seconds of planning timeout. In order to compare with previous approaches, we use Fast Downward \cite{Helmert:FD:JAIR06} in the LAMA-2011 setting~\cite{richter2011lama} to synthesize and evaluate programs using the presented compilations.

Experiments are carried out over the following generalized planning tasks. The {\em Green Block} consist of a tower of blocks where only one greenish block exists and must be collected. In {\em Fibonacci} the aim is to output the correct number in the Fibonacci sequence. In {\it Gripper} a robot has to move a number of balls from room A to room B, and in {\it List} the aim is to visit all elements of a linked list. Finally, in {\it Triangular Sum} we must calculate the triangular sum represented with the formula $y=\sum^N_{0} x$. We also introduce in this paper {\it RoboPainter} (RP), where a robot should paint, given different constraints, odd cells in a corridor (see Figure~\ref{fig:basicprograms}).

\begin{table*}
\centering
\begin{footnotesize}
\begin{tabular}{l@{\hspace*{5pt}}|cc|cc|cc|cc|}
 & & & \multicolumn{2}{c|}{\bf Only Positive} & \multicolumn{2}{c|}{\bf PN-Lite} & \multicolumn{2}{c|}{\bf PN}\\
 & $n$ & $max(T)$ & Avg. Search(s) & Found(\%) & Avg. Search(s) & Found(\%) & Avg. Search(s) & Found(\%) \\\hline
RoboPainter				&	5		&	5		&	{\bf 64.58}	&	50\%	& 140.44		& {\bf 60\%}		& 86.43		& 40\%	\\ 
Green Block		&	4		&	5		&	{\bf 45.40}	&	81.25\%	& 154.35	& 67.5\%	& 99.12		& {\bf 90\%}	\\ 
Fibonacci		&	5		&	5		&	190.14	&	25\%	& {\bf 93.96}		& {\bf 27.5\%}	& -			& 0\%	\\ 
Gripper			&	4		&	5		&	{\bf 48.19}	&	{\bf 43.75\%}	& 67.77		& 27.5\%	& 107.14	& 27.5\%	\\ 
List			&	3		&	5		&	{\bf 0.04}	&	31.25\%	& 0.07	& 27.5\%	& 0.21	& {\bf 45\%}	\\ 
T. Sum			&	3		&	5		&	143.36	&	{\bf 100\%}	& 192.13	& {\bf 100\%}	& {\bf 141.74} & {\bf 100\%}	\\ 
\end{tabular}
\end{footnotesize}
\caption{\small {\em Program synthesis with positive and negative examples}: number of program lines (n), max number of input instances (T), average search time (secs) and, percentage of found solutions (with {\em only positive} and with {\em positive and negative} examples).}
\label{tab:synthesis}
\end{table*}

\subsection{Computing programs with positive and negative examples}
For the synthesis of programs that solve the previous generalized planning tasks, we compare two versions of our compilation, {\em PN-Lite} and {\em PN}, with the results from some problems whose solutions where solved and reported as {\em ``One Procedure''} in~\citeauthor{javi-Gplanning-ICAPS16}~(\citeyear{javi-Gplanning-ICAPS16}). We use {\em PN} to denote the version with positive and negative examples that detect the three possible failures of a planning program, whereas {\em PN-Lite} is a simpler sound version that detects {\em incomplete programs} and {\em inapplicable actions} but not {\em infinite loops}.

In this experiment we have run almost 100 random configurations with at most 5 instances that could be either positive or negative (where at least one is forced to be positive, see the previous section). The idea behind this experiment is to evaluate the use of negative examples as counter-examples to prevent undesired generalizations of programs that are synthesized from small input instances. Some domains from previous approaches are simple enough that they can generalize from few positive instances, so our compilations will only add complexity to the domain, requiring extra searching time required for failure detection. 

In Table~\ref{tab:synthesis}, columns {\em PN-Lite} and {\em PN} report the obtained results when we synthesize programs that are validated on both {\em positive} and {\em negative} examples. Recall that the $P_n''$ compilation has additional fluents and actions compared to $P_n$, which imposes an extra searching time. However, including negative examples often makes it possible to synthesize programs from fewer positive examples and with fewer fluents (planning instances of smaller size) which, in general, increases the percentage of programs found. Also, the process of synthesis from few examples is a benefit in generalized planning compilations akin to few-shot learning~\cite{lake2015human,camacho2019models}.

We briefly describe the best solutions that generalize for each domain in Table~\ref{tab:synthesis}. In {\it Green Block}, we repeat the drop and unstack instructions while the green block is not hold, then we collect the holding green block. In the {\it Fibonacci} domain there are 4 variables called A, B, C and D that represent $F_n$, $n$, $F_{n-1}$ and $F_{n-2}$ respectively. The program assigns C to D, then A to C, then adds D to A and decreases B, repeating this sequence while B is different from 0. The planning program found in {\it Gripper} picks up a ball with the left hand, moves to room B, drops the ball, and goes back until no more balls are in room A. The {\it List} program visits the current node, moves to the next node and repeats the process until it reaches the tail. Finally, the program for {\it Triangular Sum} has a variable A initialized to 0 and countdown variable B that is added iteratively to A.

\begin{table*}
\centering
\begin{footnotesize}
\begin{tabular}{l|ccc|ccc|ccc|}
& $re(\Pi_\text{P})$ & $pr(\Pi_\text{P})$ & $ac(\Pi_\text{P})$ & $re(\Pi_{\text{PN-Lite}})$ & $pr(\Pi_{\text{PN-Lite}})$ & $ac(\Pi_{\text{PN-Lite}})$ & $re(\Pi_{\text{PN}})$ & $pr(\Pi_{\text{PN}})$ & $ac(\Pi_{\text{PN}})$ \\\hline
RoboPainter		&	71.74\%	& {\bf 100.00\%} & 95.19\% & 70.90\% & {\bf 100.00\%} & 94.58\% & {\bf 75.63\%} & {\bf 100.00\%} & {\bf 95.57\%} \\  
Green Block		&	68.86\% & 80.93\% &	91.48\% & 60.48\% &	75.00\%  & 88.76\% & {\bf 80.89\%} &	{\bf 88.38\%}	& {\bf 94.43\%}  \\
Fibonacci		&	17.86\% & 71.43\% &	85.47\% & {\bf 22.22\%} &	{\bf 100.00\%} & {\bf 85.97\%} & -\% & -\% & -\%	\\ 
Gripper			&	41.54\% & 85.71\% &	87.68\% & {\bf 62.38\%} &	{\bf 88.73\%} & {\bf 91.35\%} & 35.44\% & 84.88\% &	86.30\% \\ 
List			&	{\bf 8.08\%} & {\bf 72.73\%} & {\bf 81.89\%} & 5.96\% & 65.00\% & 81.00\% & 4.75\% & 47.22\% & 80.27\% \\
T. Sum			&	71.74\% & {\bf 100.00\%} & 95.19\% & {\bf 75.63\%} & {\bf 100.00\%} & {\bf 95.57\%} & 70.90\% & {\bf 100.00\%} & 94.58\% \\ 
\end{tabular}
\end{footnotesize}
\caption{\small {\em Program evaluation wrt a set of positive and negative tests} using the {\em recall}, {\em precision} and {\em accuracy} metrics. The -\% symbol refers either to value not found because of an invalid operation.}
\label{tab:validation}
\end{table*}

\subsection{Evaluating generalized plans with test sets of positive and negative examples}
{\em Negative} examples are useful for defining quantitative metrics that evaluate the coverage of generalized plans with respect to a {\em test set} of unseen examples. Given a labeled classical planning instance $P$ and a program $\Pi$:
\begin{itemize}
\item If $P$ is labeled as {\em positive} and $\Pi$ solves $P$ this means $P$ is a {\bf true positive}. Otherwise, if $\Pi$ fails to solve $P$ this means $P$ is a {\bf false positive}.
\item If $P$ is labeled as a {\em negative} example and $\Pi$ solves $P$ this means $P$ is a {\bf false negative}. Otherwise, if $\Pi$ fails to solve $P$ this means $P$ is a {\bf true negative}. 
\end{itemize}

Our notion of {\em positive} and {\em negative examples} allows us to adopt metrics from ML for the evaluation of planning solutions that generalize with respect to a {\em test set}. These metrics are more informative than simply counting the number of positive examples covered by a solution and also consider the errors made over the {\em test set}~\cite{davis2006relationship}:
\begin{itemize}
\item {\bf Precision}, $pr(\Pi)=\frac{p}{(p+p^-)}$, and {\bf Recall}, $re(\Pi)=\frac{p}{(p+n^-)}$, where $p$ is the number of {\em positive} examples solved by $\Pi$, $p^-$ the number of {\em false positives} (classical planing problems labeled as {\em negative} that are solved by $\Pi$) and $n^-$ is the number of {\em false negatives} (instances labeled as positive examples that cannot be solved by $\Pi$). 
\item {\bf Accuracy} is a more informed metric frequently used in ML, $ac(\Pi)=\frac{p+n}{p+n+p^-+n^-}$. In our case, $n$ represents the number of {\em negative} examples unsolved by the program $\Pi$. 
\end{itemize}

For this experiment we considered the planning program as given, i.e.~the computed programs in the previous synthesis experiment. Then we compile a set of positive and negative instances but include the planning program in the initial state instead of having {\em empty} lines (as in the $P_n$ and $P_n'$ compilations). The execution of the computed programs on these instances must solve the positive instances while failing to solve the negative instances, verifying plan failure due to a failed condition or the detection of an infinite loop, as explained in the previous section. 

We have used a framework for validating planning programs that reports {\em success} or specifies the {\em source of failure} when executing the program for each randomly generated planning instance. The results are shown in Table~\ref{tab:validation} where we report the {\em precision}, {\em accuracy} and {\em recall} of programs synthesized using {\em only positive examples}, and programs synthesized using the {\em positive and negative examples}. The list domain is the only case where positive examples yield to a better accuracy, while the rest of domains using positive and negatives improves only positives in all metrics.

\section{Conclusion}
{\em Generalized planning} provides an interesting framework to bridge the gap between ML and {\em AI planning}~\cite{geffner2018:mbased-mfree:ijcai18}. On the one hand {\em generalized planning} follows a model based approach with declarative definitions of actions and goals, as {\em AI planning}. On the other hand {\em generalized planning}, as inductive ML, computes solutions that generalize over a set of input examples. {\em Generalized planning} has however little work dedicated to the assessment of the generality of plans beyond the given input planning tasks (positive examples only). ML however, has a long tradition on the empirical assessment of the generality of solutions. The fact that our compilation identifies the source of failures of program execution on a particular planning instance allows us to introduce negative examples and to bring the evaluation machinery from ML to define evaluation metrics that empirically assess the generality of plans beyond the given input planning tasks, e.g. using {\em test sets}.

{\em Model checking}~\cite{clarke:ModelC:book99} provides effective solvers for automatically verifying correctness properties for diverse finite-state systems. More precisely when actions have non-deterministic effects, program validation becomes complex since it requires proving that all the possible program executions reach the goals. In such a scenario, {\em model checking}~\cite{clarke:ModelC:book99} and {\em non-deterministic planning}, like POMDP planning, are definitely more suitable approaches for plan validation~\cite{hoffmann:pentesting:ICAPS2015}. An interesting research direction is to study how to leverage {\em model checking} techniques for the synthesis of generalized planning form both positive and negative examples. {\em Plan constraints} are also a compact way of expressing {\em negative} information for planning and reduce the space of possible solution plans. Plan constraints have already been introduced to different planning models using the LTL formalism~\cite{baier:ltl:AI09,Haslum:LTLf:ECAI10,Nir:LTL:IJCAI13,camacho2017non}.

\section*{Acknowledgments}
The research leading to these results has received funding from the European Union’s Horizon 2020
research and innovation programme under grant agreement no. 731761, IMAGINE; and it is partially supported by grant TIN-2015-67959 and the Maria de Maeztu Units of Excellence Programme MDM-2015-0502, MEC, Spain. Sergio Jim\'enez is supported by the {\it Ramon y Cajal} program, RYC-2015-18009, the Spanish MINECO project TIN2017-88476-C2-1-R, and the {\em generalitat valenciana} project GV/2019/082. Anders Jonsson is partially supported by the Spanish grants TIN2015-67959 and PCIN-2017-082.

\bibliographystyle{aaai}
\bibliography{negative-examples}

\end{document}